\documentclass[letterpaper, 10 pt, conference]{ieeeconf}  

\IEEEoverridecommandlockouts                              

\usepackage{graphics} 
\usepackage{amsmath} 
\usepackage{amssymb}  
\usepackage{graphicx} 
\usepackage{booktabs} 
\usepackage{hyperref}
\usepackage{bbm}

\title{\LARGE \bf
Navigating Robots from Unknown and Uncalibrated Exocentric Cameras via Map Estimation
}

\author{Raul Steinmetz$^{1,2}$, Raphael Druon$^{2}$, Yusuke Yoshiyasu$^{2}$, Akihisa Ohya$^{1}$, and Ayanori Yorozu$^{1}$%
\thanks{%
$^{1}$Authors are with the Intelligent Robot Laboratory, University of Tsukuba, Japan.
\texttt{raulsteinmetz08@gmail.com, \{ohya, yorozu\}@cs.tsukuba.ac.jp}\newline
$^{2}$Authors are with the Computer Vision Research Team, Artificial Intelligence Research Center (AIRC), National Institute of Advanced Industrial Science and Technology (AIST), Japan.
\texttt{\{raphael.druon, yusuke-yoshiyasu\}@aist.go.jp}\newline
\textbullet~This work was supported by the Ministry of Education, Culture, Sports, Science and Technology (MEXT), Japan.\newline
\textbullet~Part of the experiments were run on the AI Bridging Cloud Infrastructure (ABCI 3.0) provided by AIST.\newline
\textbullet~Code and data: \url{https://github.com/raulsteinmetz/romux}
}%
}

\begin{document}

\maketitle
\thispagestyle{empty}
\pagestyle{empty}

\begin{abstract}
Exocentric cameras, mounted on walls or ceilings in indoor environments, offer wider vantage points and are less easily occluded than robot-mounted cameras, making them an appealing alternative for perception in robot navigation algorithms. Previous methods that use such cameras, however, require their pose and intrinsics to be known before deployment, demanding calibration that is time-consuming, requires specialized personnel, and must be repeated whenever a camera moves or a new room is navigated. We propose Robot-centric Obstacle Maps from Unknown and uncalibrated eXocentric cameras (ROMUX), an agent that navigates a robot toward point goals while avoiding obstacles using exocentric cameras that are both uncalibrated, meaning the system does not require their pose or intrinsic information as input, and unknown, meaning it is not fit to a specific camera during training and does not require pre-deployment fine-tuning. We train a cross-view transformer on large-scale simulation data to predict metric robot-centric obstacle maps from exocentric images, without conditioning image tokens on camera information. A small policy trained with Proximal Policy Optimization (PPO) then navigates on the predicted maps toward a goal coordinate relative to the robot. We benchmark our agent in simulation with cameras randomized per episode, and it matches the success rates of a PPO policy that navigates from a robot-mounted camera. Further, ROMUX works with one camera or multiple cameras with partial views and has been transferred zero-shot  to a real robot using a camera at varying positions and orientations across various room layouts.
\end{abstract}

\section{INTRODUCTION}\label{intro}

Navigating towards a goal while avoiding collisions with obstacles is an essential capability for many autonomous service robots. Recent trends in robotics research have explored alternatives to classical mapping and planning algorithms, towards agents that learn to navigate from experience and perceive their environment using raw images. Deep reinforcement learning (DRL) has emerged as a central methodology for learning such navigation policies. Given enough interactions with a simulated environment, DRL produces neural network policies capable of navigating unknown environments with state-of-the-art success rates \cite{wijmans2019dd}.

However, most navigation policies in this context use cameras mounted on the robot frame for perception, while few works have explored the usage of exocentric cameras. Exocentric cameras, such as monitoring cameras mounted on walls and ceilings in indoor environments, offer several advantages for indoor navigation, such as: wider coverage of the environment in a single observation and less risk of heavy occlusion than robot-mounted sensors; no drain on robot battery or payload; and availability in many of the environments where service robots already operate, enabling their use at little extra cost.

Previous work on navigation from exocentric cameras requires cameras to be known in advance, either through classical calibration that recovers camera pose and intrinsics \cite{bultmann2023external, kim2024cctv} or through pre-deployment learning stages that fit perception models to each camera view \cite{robinson2023robot, robinson2026multi}. Both assumptions are limiting, as classical calibration demands specialized personnel and time and must be repeated whenever a camera moves or a new room is navigated, and fitting models to specific views likewise requires retraining for every new room and camera placement.

\begin{figure}[t]
\vspace{7pt}
\centering
\includegraphics[width=\columnwidth]{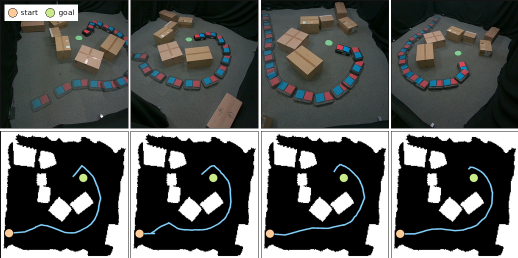}
\caption{ROMUX trajectories on the real robot. Each column is one episode under a different camera position on the same obstacle layout.}
\label{fig:real-exp}
\vspace{-21pt}
\end{figure}

We propose the problem of navigating robots toward a point-goal while avoiding collisions with obstacles, using images from exocentric cameras that are both \emph{uncalibrated}, meaning their pose and intrinsics are not provided to the system, and \emph{unknown}, meaning the agent is not fitted to any specific camera during training. To enforce these constraints, we randomize camera pose and intrinsics at every evaluation episode. We show that Proximal Policy Optimization (PPO)~\cite{schulman2017proximal}, which achieves near-perfect point-goal navigation from robot-mounted camera images~\cite{wijmans2019dd}, fails to learn obstacle avoidance from exocentric cameras in our setup, even when a fixed camera is used throughout training and evaluation. We hypothesize that this occurs because obstacle avoidance requires mapping obstacles from the exocentric image into the robot frame, and PPO does not recover this transformation within our setup and training budget.

To navigate robots under this setup, we propose Robot-centric Obstacle Maps from Unknown and uncalibrated eXocentric cameras (ROMUX), an agent architecture that decouples the aforementioned perspective transformation from the reinforcement learning objective and formulates it as a supervised learning problem. We collect diverse pairs of exocentric camera images and obstacle maps at scale in simulation, randomizing camera pose and intrinsics, and environment layout per data point, and train a cross-view transformer-like neural network \cite{zhou2022cross} on top of image features from a fine-tuned DINOv2 \cite{oquab2024dinov2} to learn the mapping between the two.

\begin{figure*}[t]
\centering
\includegraphics[width=0.75\textwidth]{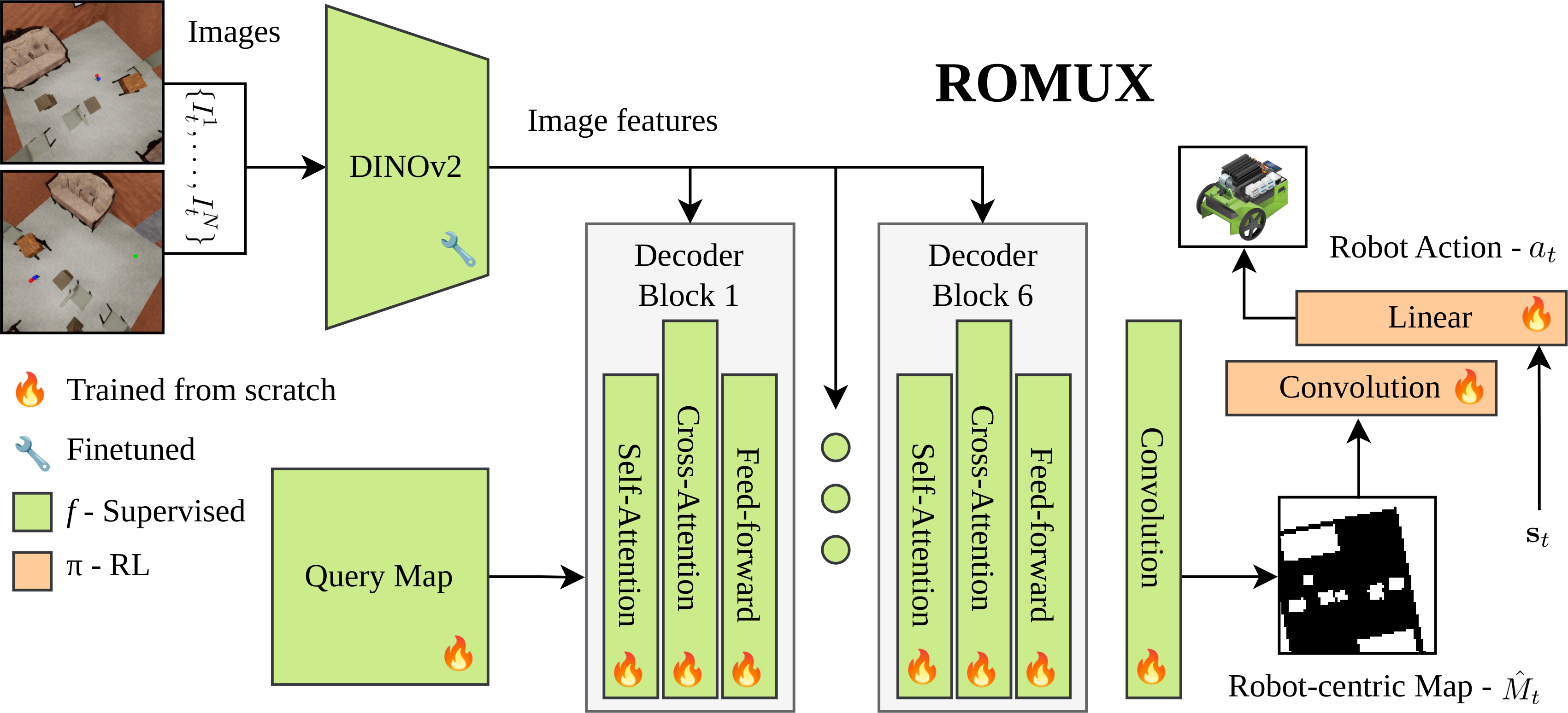}
\caption{Simplified diagram of the ROMUX agent architecture. }
\label{fig:architecture}
\vspace{-15pt}
\end{figure*}

The map-estimation module predicts a robot-centric binary obstacle map without receiving camera pose or intrinsics, and a small policy trained with PPO navigates from the estimated map and a goal coordinate relative to the robot.  We only require that the robot and its heading be visually identifiable in the image; in our setup, we use two colored markers on top of the robot so that this information can be recovered from the low-resolution image. The two stages are trained separately and do not share gradients, so the map-estimation module can be retrained for a new domain while the policy is recycled. The agent scales to multiple cameras with partial views and has been  transferred zero-shot to a real robot, where a camera is placed at different positions and orientations and the obstacle layout varies across trials (Figure~\ref{fig:real-exp}). Figure~\ref{fig:architecture} shows a diagram of the architecture. Our main contributions are the following:

\begin{itemize}
\item A new problem for the robot navigation literature, where an algorithm must navigate a robot using unknown and uncalibrated exocentric cameras;
\item The ROMUX architecture, which can navigate the proposed setup.
\end{itemize}

\section{RELATED WORK}\label{related}

Anderson et al.~\cite{anderson2018evaluation} define point-goal navigation as the task of navigating an agent towards a goal specified as a coordinate relative to the agent. We adopt that definition but add the requirement that the robot must not collide with any obstacle on its path to the goal. On point-goal navigation, Wijmans et al.~\cite{wijmans2019dd} show that PPO, when distributed and scaled to 2.5 billion training steps, can achieve near-perfect point-goal navigation in indoor environments, although obstacle avoidance is not considered. That work uses robot-mounted sensors, while we navigate under exocentric cameras.

Kim et al.~\cite{kim2024cctv} use exocentric CCTV cameras to detect and track human movement in crowded indoor environments, project their image detections onto a known global occupancy map via a homography transformation, and plan a global path that a deep reinforcement learning local planner then follows. Bultmann et al.~\cite{bultmann2023external} estimate the pose of a robot with respect to a network of exocentric cameras by detecting keypoints of the robot in each image and minimizing the multi-view projection error, using that information to correct the localization of a robot that still navigates using its onboard sensors. Both works require calibrating the pose and intrinsics of every camera beforehand. Recalibration is then needed whenever a camera moves or the robot must navigate a new room. Instead, we estimate robot-centric obstacle maps directly from images captured by cameras with unknown pose and intrinsics.

Most similar to this work, Robinson et al.~\cite{robinson2023robot, robinson2026multi} navigate robots using exocentric camera networks without requiring explicit knowledge of each camera's pose and intrinsics. Their system, however, still requires a calibration stage that involves training a robot detector, segmenting the navigable region, and tuning a navigation controller on each camera. They also require up to 30 minutes of random robot walks to discover handover regions between overlapping camera views. This calibration stage must be repeated whenever a camera is added or moved, and the drivable region it produces is static, so it does not account for obstacles that change position. In contrast, we estimate a robot-centric obstacle map at every timestep, naturally handling changing environments, and require no calibration or fitting models to specific cameras.

Finally, our architecture draws on neural mapping and planning navigation algorithms. Gupta et al.~\cite{gupta2017cognitive} accumulate views from a robot-mounted camera into a top-down belief map centered around the robot frame and run a differentiable planner on top of it. A parallel line of work exists in autonomous driving, where bird's-eye-view models predict map-view representations of a scene from cameras mounted on the vehicle. Zhou et al.~\cite{zhou2022cross} achieve this with a cross-view transformer, in which embeddings derived from the intrinsics and extrinsics of each camera guide attention from image features towards the map-view representation. We adopt the cross-view transformer as our map-estimation module but remove the per-camera embeddings because the camera configurations are unknown. Pan et al.~\cite{pan2020cross} predict a top-down semantic map from first-person views without using camera parameters. However, their cameras form a fixed rig carried by the agent. These works estimate maps from robot-mounted or vehicle-mounted cameras, while we estimate them from exocentric cameras.

\section{METHODOLOGY}\label{method}

The fundamental difficulty of navigating from exocentric cameras with unknown configuration is that no geometric projection is available to map what the cameras see into the robot frame, suggesting the use of neural networks to learn this mapping instead. In Section~\ref{experiments}, we show that an end-to-end reinforcement learning approach, such as PPO, does not implicitly learn such a transformation in our setup, suggesting that it should be learned explicitly.

\subsection{Problem Formulation}

We consider a point-goal navigation task in which a differential-drive robot must reach a goal position without colliding with obstacles, using images from exocentric cameras with unknown pose and intrinsics. The robot carries two colored markers on its chassis, one at the front and one at the rear, making it visually identifiable and allowing its heading to be recovered from an exocentric view at low image resolution. At each timestep $t$, the agent receives a state vector $\mathbf{s}_t = (\cos\theta_t, \sin\theta_t, d_t, v_t, \omega_t)$, where $\theta_t$ and $d_t$ are the bearing and distance to the goal in the robot frame, and $v_t$ and $\omega_t$ are the current forward and angular velocities. 

 Alongside $\mathbf{s}_t$, the system receives a set of $N$ exocentric camera images $\{I^1_t, \dots, I^N_t\}$ from cameras placed in the environment. In our setup, the pose and intrinsics of each camera are randomized at each episode and never provided to the system. The robot selects an action $a_t \in \{\textit{forward}, \textit{left}, \textit{right}\}$, each applying a fixed linear or angular velocity for one step. An episode terminates when the robot reaches within a fixed radius of the goal (success), collides with an obstacle or wall (failure), or exceeds the maximum episode length (time limit).

\subsection{Agent Architecture}

The ROMUX agent decouples navigation into two stages. A map-estimation module $f$ maps the exocentric images into a metric robot-centric obstacle map $\hat{M}_t = f(I^1_t, \dots, I^N_t)$, where $\hat{M}_t \in \{0,1\}^{H \times H}$ is a binary grid centered on the robot and aligned with its heading. A policy $\pi$ then selects actions from the predicted map and the state vector, $a_t \sim \pi(\hat{M}_t, \mathbf{s}_t)$. The two stages are trained separately and do not share gradients. The map-estimation module is trained with supervised learning on ground-truth robot-centric obstacle maps $M_t$ obtained from simulation, while the policy is trained with PPO on maps estimated by $f$, also in simulation. Because the two stages are connected through a binary map representation, either stage can be replaced independently of the other.

\subsection{Map Estimation and Policy Training}

The map-estimation module $f$ must predict a metric, robot-centric obstacle map aligned with the robot heading from a set of images with unknown camera pose and intrinsics, a problem with no known trivial analytic solution. We build $f$ on the cross-view transformer architecture~\cite{zhou2022cross} but remove the camera embeddings that the original neural network conditions on, since we have no access to camera information. Instead, the map-estimation module relies solely on image content to infer obstacle positions and orientations relative to the robot, which we show is possible.

Each image is encoded by a DINOv2 ViT-S/14~\cite{oquab2024dinov2} backbone into a sequence of patch tokens.  The patch tokens from all $N$ cameras are projected into a shared embedding space and summed with a learnable patch-position embedding shared across views, making the fusion permutation-invariant with respect to camera ordering. A $32 \times 32$ grid of learnable map queries, each corresponding to a $2 \times 2$ block of cells in the output grid, attends to the concatenated image tokens through a six-layer transformer decoder with pre-norm self-attention, cross-attention, and a feed-forward network. The decoded queries are reshaped into a spatial grid and passed through a convolutional head that upsamples them to $H \times H = 64 \times 64$ and outputs per-cell occupancy logits.

The map-estimation module is trained on ground-truth robot-centric obstacle maps $M_t$. Each cell corresponds to a fixed physical extent, and cells outside the room walls are labeled as occupied. The map covers a fixed metric window around the robot, which is essential because a global map would require higher dimensionality as the environment size increases, whereas a local map scales to arbitrary environment sizes. However, it comes at the cost of losing information about objects that can be seen in the image but are not around the robot. Let $\hat{m}_{ij}$ be the occupancy probability the module assigns to cell $(i,j)$, obtained by applying a sigmoid to the predicted logit, and $m_{ij} \in \{0, 1\}$ be the ground-truth occupancy label. The map-estimation module is trained by minimizing a combination of binary cross-entropy and Dice loss~\cite{milletari2016v} over the $H \times H$ cells,

\begin{equation}
\mathcal{L}_f = \alpha \, \mathcal{L}_{\text{BCE}} + (1 - \alpha) \, \mathcal{L}_{\text{Dice}},
\end{equation}

\noindent where

\begin{equation}
\mathcal{L}_{\text{BCE}} = -\frac{1}{H^2} \sum_{(i,j)} \left[ m_{ij} \log \hat{m}_{ij} + (1 - m_{ij}) \log (1 - \hat{m}_{ij}) \right],
\end{equation}

\noindent and

\begin{equation}
\mathcal{L}_{\text{Dice}} = 1 - \frac{2 \sum_{(i,j)} \hat{m}_{ij} \, m_{ij} + 1}{\sum_{(i,j)} \hat{m}_{ij} + \sum_{(i,j)} m_{ij} + 1}.
\end{equation}

During training, camera images are augmented with random color jitter and additive pixel noise, applied independently to each view, to avoid overfitting to patterns and enable sim-to-real transfer through domain randomization~\cite{tobin2017domain}. The DINOv2 backbone is frozen for the first epochs and then fine-tuned with a lower learning rate, while the decoder and convolutional head are trained from scratch with a linear warmup followed by cosine annealing.

The navigation policy $\pi$ is trained with PPO~\cite{schulman2017proximal} on the maps predicted by $f$. The policy receives the obstacle map $\hat{M}_t$ as a one-channel image and the state vector $\mathbf{s}_t$ as a separate input. Convolutional layers process the map into a vector embedding, which is concatenated with $\mathbf{s}_t$ and passed to separate policy and value MLPs. The reward at each timestep is

\begin{equation}
r_t = \lambda_p \, (d_{t-1} - d_t) + \lambda_s \, \mathbbm{1}[\text{success}] - \lambda_c \, \mathbbm{1}[\text{collision}] - \lambda_{\tau},
\end{equation}

\noindent where $d_t$ is the distance to the goal, $\lambda_p$ scales the progress term, $\lambda_s$ and $\lambda_c$ are the success and collision rewards, and $\lambda_{\tau}$ is a per-step penalty that encourages shorter trajectories. The policy is trained to maximize the expected discounted return

\begin{equation}
J(\pi) = \mathbb{E}_{\pi} \left[ \sum_{t=0}^{T} \gamma^t \, r_t \right],
\end{equation}

\noindent where $\gamma$ is the discount factor and $T$ is the episode length. PPO maximizes this objective through a clipped surrogate loss that bounds the ratio between the current and sampling policies and weights it by generalized advantage estimates~\cite{schulman2015high}. Hyperparameters are reported in Section~\ref{experiments}.

At deployment, the map-estimation module processes the exocentric images into a binary robot-centric obstacle map at each timestep by thresholding occupancy probabilities at 0.5, and the policy selects an action from that map and the robot state. We hypothesize that when trained at scale with cameras randomized for each snapshot, the map-estimation module learns to locate the robot within the set of images and derive a map of the surrounding obstacles relative to it.

\subsection{Simulation Environment}

We build our simulation environments in NVIDIA Isaac Lab~\cite{mittal2025isaac}. Each environment consists of a rectangular room whose dimensions are sampled uniformly per episode from a configured range. The room contains rectangular cardboard obstacles whose quantities, sizes, positions, and orientations are randomized within configured ranges. The robot and goal poses are sampled within the room bounds. In $70\%$ of episodes, the robot is placed near one of the four corners of the room, and the goal is set as the farthest from the robot among five candidates drawn uniformly over the room, so these episodes require longer traverses. In the remaining $30\%$, the robot is placed uniformly, and the goal is the first candidate drawn, where goal candidates that overlap the robot are discarded. The robot heading is sampled uniformly in both cases. Obstacles are then placed sequentially, rejecting any candidate that overlaps a previously placed obstacle, the robot, or the goal. A breadth-first search on the global obstacle map of the room, dilated by the robot's radius, verifies that a traversable path connects the robot to the goal, and the entire placement is resampled if no such path exists.

Each exocentric camera is placed by sampling its position from a configured spatial range and orienting it towards the room center with a random offset. A camera's field of view varies with a zoom factor $\zeta$ drawn from a configured range, applied to the focal length $F$ and the horizontal aperture $W$ as $F' = F/\sqrt{\zeta}$ and $W' = W\sqrt{\zeta}$. The horizontal field of view is then $2\arctan\!\left(W'/2F'\right) = 2\arctan\!\left(\zeta\,W/2F\right)$, so $\zeta$ scales the tangent of the half-angle directly and, with it, the extent of the scene the camera covers at a fixed distance. Wall, floor, and obstacle textures are redrawn from a pool of photorealistic materials at every reset. All randomizations, geometric and visual, are resampled every episode, so that different episodes do not share the same obstacle layout, camera configuration, or appearance. Training data for the map-estimation module is collected by resetting the simulator and saving a snapshot of each environment, so that every snapshot is an independent scene.

\section{EVALUATION}\label{experiments}

We evaluate ROMUX against PPO baselines in the \textit{cardboard rooms}, small rooms where obstacles, camera pose, and intrinsics are randomized at every episode (Section~\ref{first-set-of-experiments}). We then evaluate the map-estimation module in environments with photorealistic furniture (Section~\ref{photorealistic-experiment}) and in a large room observed by four cameras with partial views (Section~\ref{four-camera-experiment}). Section~\ref{occupancy-iou} compares the quality of the predicted maps across all settings, and Section~\ref{real-robot-experiment} reports the real robot experiment.

\subsection{Simulation Setup}

All experiments use different versions of the environment described in Section~\ref{method}. Here, we describe the settings used for the \textit{cardboard rooms} (Subsection~\ref{first-set-of-experiments}), and for other experiments, we will specify the changes to this setup in their own subsections. The room width is fixed at 6 m, and the depth is sampled uniformly from 3 to 6 m. Each episode attempts to place 6 to 12 rectangular cardboard obstacles with randomized sizes and orientations, dropping any candidate that cannot be fitted without overlap after 5 tries. Obstacle footprints are sampled with side lengths in $[0.2, 2.5]$ m by $[0.2, 1.5]$ m and heights in $[0.2, 0.5]$ m, and the total obstacle area is capped at $80$\% of the room floor. 

The robot operates at 5 Hz, with discrete actions either applying a forward speed of 0.30 m/s or turning by 10 degrees. An episode succeeds when the robot reaches within 0.2 m of the goal and times out after 250 steps. The reward coefficients are $\lambda_p = 4$, $\lambda_s = 100$, $\lambda_c = 10$, and $\lambda_\tau = 0.01$.

Camera images are rendered at $256 \times 256$ pixels for exocentric configurations and at $84 \times 84$ pixels for the robot-mounted camera. Exocentric configurations use a higher resolution because the robot must remain visible from a distance. In the one-camera configuration for ROMUX, the camera position is sampled with $x \in [-2, 2]$ m, $y \in [-0.8, 0.8]$ m, and height $z \in [2.7, 2.9]$ m, oriented towards the room center with a look-at jitter of $\pm 0.1$ m, focal length 16 mm, horizontal aperture 45 mm, and zoom factor in $[0.9, 1.0]$. The two-camera configuration places each camera in one half of the room along the $x$-axis, with $x \in [0.0, 1.2]$ m for one camera and $x \in [-1.2, 0.0]$ m for the other, $y \in [-0.8, 0.8]$ m, and height $z \in [2.2, 2.6]$ m. The first camera points at $(1.0, 0.0)$ m and the second at $(-1.0, 0.0)$ m relative to the room center, both with a look-at jitter of $\pm 0.1$ m, focal length 9 mm, horizontal aperture 24 mm, and zoom factor in $[0.9, 1.0]$. These ranges were selected so that, in the one-camera configuration, most of the room is visible from every sampled pose. In contrast, in the two-camera configuration, a camera rarely sees the whole environment, but the two partial views combined do. In every multi-camera experiment, the input images are randomly shuffled, so the agent cannot tie a specific camera to a region of the room.

The fixed exocentric camera of the end-to-end PPO baseline is instead held at a fixed pose across all training and evaluation episodes, at $(0.5, 0.0, 3.2)$ m, pointing at the room center with no look-at jitter and no zoom randomization, with focal length 19 mm and horizontal aperture 40 mm.

The ROMUX map-estimation module has 29.4 M parameters and is trained separately for 30 epochs on 80,000 independently collected snapshots, with a batch size of 128. The obstacle map spans a $6 \times 6$ m window at $64 \times 64$ resolution, and camera images are resized to $224 \times 224$ for the DINOv2 backbone. The decoder learning rate is $2 \times 10^{-4}$ and the backbone learning rate is $4 \times 10^{-5}$ after unfreezing at epoch 2. The map-estimation loss weights BCE and Dice equally at $\alpha = 0.5$. During training, each image is augmented with color jitter (brightness, contrast, and saturation varied by $\pm 0.4$, hue by $\pm 0.1$) applied with probability 0.8, and additive pixel noise with standard deviation sampled uniformly from $[0, 15]$ on an 8-bit scale.

All policies use the same PPO hyperparameters, with a learning rate linearly decayed from $3 \times 10^{-4}$ to $5 \times 10^{-5}$, clip range from $0.2$ to $0.1$, entropy coefficient from $0.05$ to $0.005$, discount factor $\gamma = 0.995$, GAE parameter $\lambda_{\text{GAE}} = 0.95$, KL divergence target of $0.03$, batch size of 4096, and 4 optimization epochs per update. ROMUX and all PPO baselines use the same policy network, in which each image input is encoded by three convolutional layers with 32, 64, and 64 filters, kernel sizes of 8, 4, and 3, and strides of 4, 2, and 1, followed by a linear layer of 256 units~\cite{mnih2015human}. This embedding is concatenated with the five-dimensional state vector and passed to separate policy and value MLPs with two hidden layers of 128 units and Tanh activations. In contrast, the baseline with no image input passes the state vector directly to the MLPs. Each configuration is trained for 20 million timesteps on 128 parallel environments with three independent seeds.

Evaluation runs 10,000 episodes per seed, with room layouts, obstacles, and camera configurations (for ROMUX only) randomized for each episode. We report the mean and 95\% confidence interval of the success, collision, and time-limit rates across the three seeds. Every interval in this paper is a Student's $t$ interval over the per-seed values, with a half-width of $4.30 \, s / \sqrt{3}$, where $s$ is the sample standard deviation across seeds.

We acknowledge that ROMUX receives additional training data during the supervised map-estimation stage, thereby increasing the total training budget compared with the PPO baselines. Extending training to 50 million timesteps for the exocentric PPO baseline in the \textit{cardboard rooms} experiment of Section~\ref{first-set-of-experiments}, however, results in $24.55 \pm 2.81$\% success, compared to $24.64 \pm 3.15$\% at 20 million timesteps, so its failure to avoid obstacles is likely not a matter of training budget.

\subsection{Benchmarking ROMUX against PPO baselines}\label{first-set-of-experiments}

We compare ROMUX against four baselines. \textit{PPO with no image input} receives only the state vector, a lower bound on what is achievable without obstacle information. \textit{PPO with fixed exocentric camera} uses a fixed top-down camera, testing whether PPO alone can avoid obstacles from an exocentric view. \textit{PPO with robot-mounted camera} navigates from a camera on the robot chassis, the standard setup in prior work. \textit{PPO with ground-truth obstacle map} navigates on the true robot-centric map from the simulator, an upper bound if map estimation were perfect.

\begin{table}[t]
\centering
\caption{Navigation results in the cardboard rooms}
\label{tab:cardboard}
\begin{tabular}{lccc}
\toprule
Agent & Success & Collision & Time Limit \\
\midrule
PPO No Image & $25.80 \pm 2.55$ & $74.19 \pm 2.61$ & $0.01 \pm 0.06$ \\
PPO Fixed Exo & $24.64 \pm 3.15$ & $69.61 \pm 15.62$ & $5.75 \pm 12.94$ \\
PPO Robot Cam & $83.86 \pm 4.66$ & $5.97 \pm 6.28$ & $10.17 \pm 5.50$ \\
\midrule
ROMUX 1-cam & $85.04 \pm 2.42$ & $9.87 \pm 0.72$ & $5.09 \pm 1.72$ \\
ROMUX 2-cam & $\mathbf{89.56 \pm 2.51}$ & $6.13 \pm 2.38$ & $4.31 \pm 1.22$ \\
\midrule
PPO GT Map & $94.27 \pm 3.46$ & $2.32 \pm 1.38$ & $3.40 \pm 2.47$ \\
\bottomrule
\end{tabular}
\vspace{-14pt}
\end{table}

Table~\ref{tab:cardboard} presents the navigation results. PPO with no image input reaches $25.80 \pm 2.55$\% success, moving blindly towards the goal and colliding in most episodes. PPO with a fixed exocentric camera reaches $24.64 \pm 3.15$\%, within the interval of the no-image baseline. This suggests that, under our training budget, PPO does not learn to avoid obstacles from exocentric images even when the camera is held fixed across training and evaluation, and that it does not recover from the image where obstacles are relative to the robot. PPO with a robot-mounted camera succeeds in $83.86 \pm 4.66$\% of episodes, as expected for the default interface in prior work.

ROMUX with one randomized exocentric camera reaches $85.04 \pm 2.42$\% success, matching the robot-mounted baseline even though the camera pose and intrinsics change every episode. A second camera raises success to $89.56 \pm 2.51$\%, likely due to the increased map precision, above the robot-mounted configuration, although the intervals still overlap at three seeds. The ground-truth map sets the upper bound at $94.27 \pm 3.46$\%. The gap between ROMUX and the last PPO baseline likely comes from map-estimation errors combined with a policy that has learned unsafe behaviors, such as passing too close to obstacles.

\subsection{Photorealistic Furniture}\label{photorealistic-experiment}

The previous experiments use rectangular cardboard obstacles with simple shapes and textures. To evaluate whether ROMUX scales to visually complex, photorealistic scenes, we replace the cardboard obstacles with photorealistic 3D furniture assets such as sofas, tables, chairs, bookshelves, and appliances and evaluate the two-camera configuration.

In this setting, we retrain the map-estimation module while keeping the PPO policy from the \textit{cardboard rooms}, also testing the decoupling between the two modules. To train it, we use 180,000 snapshots collected from 180 room layouts containing 103 different furniture assets. In the training data, the furniture is placed at random within the room to maximize image variety. We train the map-estimation module using the same procedure as before. We pair the three PPO policies trained for the two-camera configuration in the previous experiment with three retrained map-estimation module seeds to form the three evaluation runs.

We evaluate ROMUX on two obstacle sets. The obstacle layouts used for navigation were not seen by the map-estimation module during training, and in all of them the furniture is randomized within human-tuned ranges per obstacle that keep the arrangements realistic, rather than the uniform placement used to collect the training snapshots. The \textit{known-obstacle} evaluation uses 2,000 episodes across 20 layouts whose furniture items were seen during map-estimation training. The \textit{unknown-obstacle} evaluation uses 400 episodes across 4 layouts containing 13 furniture assets held out from the map-estimation training set.

\begin{table}[t]
\centering
\caption{Navigation results with photorealistic furniture}
\label{tab:photoreal}
\begin{tabular}{lccc}
\toprule
Evaluation & Success & Collision & Time Limit \\
\midrule
Known obstacles & $86.40 \pm 2.94$ & $8.92 \pm 4.82$ & $4.68 \pm 2.17$ \\
Unknown obstacles & $74.83 \pm 6.70$ & $17.83 \pm 6.87$ & $7.33 \pm 1.90$ \\
\bottomrule
\end{tabular}
\vspace{-14pt}
\end{table}

Table~\ref{tab:photoreal} reports the results averaged over three seeds. On the known obstacle set, ROMUX achieves $86.40 \pm 2.94$\% success, close to the $89.56 \pm 2.51$\% obtained on cardboard obstacles in the same two-camera configuration (Table~\ref{tab:cardboard}), even though the PPO policy was never exposed to photorealistic scenes during training. The retrained map-estimation module produces maps similar enough that the policy transfers with little degradation. On the unknown obstacle set, the success rate drops to $74.83 \pm 6.70$\% and collisions rise to $17.83 \pm 6.87$\%, since held-out furniture has appearances absent from the map-estimation training data. The agent still succeeds in roughly three out of four episodes, so the map-estimation module partially generalizes to furniture assets it has never seen. 

\subsection{Scaling to a Larger Room with Four Cameras}\label{four-camera-experiment}

\begin{figure}[t]
\centering
\includegraphics[width=\columnwidth]{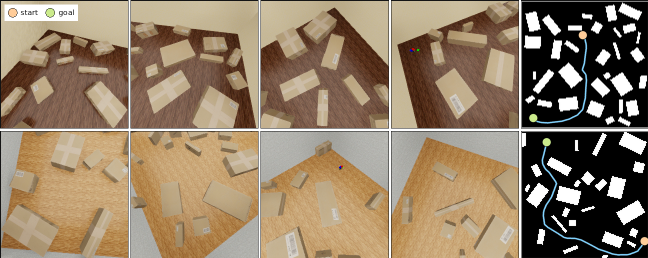}
\caption{ROMUX trajectories in the $12 \times 12$~m rooms. Each row shows one episode, with the final image for each of the four cameras, and the robot's trajectory overlaid on the ground-truth global obstacle map.}
\label{fig:4cam}
\vspace{-14pt}
\end{figure}

To test whether ROMUX scales to larger environments, we return to cardboard obstacles and increase the room to $12 \times 12$~m. We increase the number of obstacles proportionally to 24--48 per episode. We place four cameras, one per quadrant. Each reuses the one-camera configuration of the \textit{cardboard rooms}, with position, orientation, and intrinsics drawn from the same ranges, but centered on its own $6 \times 6$~m quadrant rather than on the room. Episodes time out after 500 steps instead of 250, since the goal can now be twice as far away. The obstacle map maintains the same $6 \times 6$~m extent and $64 \times 64$ resolution as in the \textit{cardboard rooms}, allowing the policy to be reused.

The map-estimation module is retrained on 160,000 snapshots from this setting, following the same training procedure with a batch size of 96, and evaluation runs 10,000 episodes per seed over three seeds. ROMUX achieves $70.21 \pm 4.98$\% success, $17.84 \pm 2.42$\% collision, and $11.95 \pm 3.43$\% time-limit rate. The map estimation IoU for the obstacles falls between the one- and two-camera cardboard IoU (Section~\ref{occupancy-iou}), so the drop in success rate likely reflects a policy that is out of distribution, with goal distances and obstacle layouts far from those seen during training. Figure~\ref{fig:4cam} shows two example trajectories.

\subsection{Obstacle Map Quality}\label{occupancy-iou}

We measure map quality with the intersection-over-union (IoU) between the predicted and ground-truth robot-centric obstacle maps, on held-out evaluation splits. For the \textit{cardboard rooms} and the $12 \times 12$~m room, the split is drawn from the same randomization ranges used for training. For the \textit{photorealistic rooms}, we report IoU separately on the known- and unknown-obstacle layout sets used for navigation in Section~\ref{photorealistic-experiment}. We report the mean and 95\% confidence interval over three seeds.

\begin{figure}[t]
\centering
\includegraphics[width=\columnwidth]{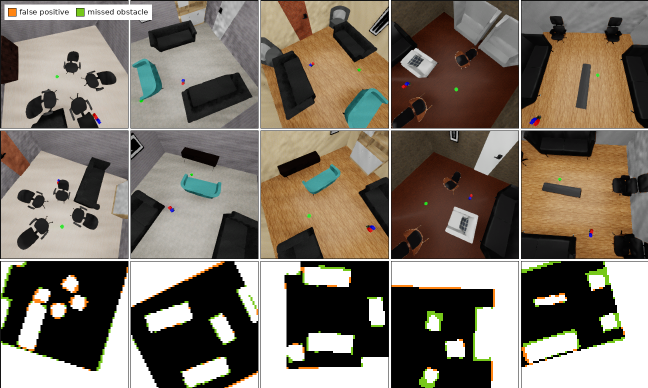}
\caption{ROMUX map estimation examples for unknown furniture in the photorealistic environment. Each column shows the two raw camera images and map prediction errors. False positives are orange and false negatives are green (for obstacles).}
\label{fig:map}
\vspace{-14pt}
\end{figure}

Because the robot-centric window is often dominated by wall cells that are trivial to predict, we report two IoU statistics, one computed over all occupied cells and another restricted to interior obstacle cells. The second is a fair basis for comparison across settings, since the $12 \times 12$~m room produces snapshots in which the robot-centric map is dominated by wall cells less often than the smaller rooms. 

\begin{table}[t]
\centering
\caption{Robot-centric obstacle map IoU (\%)}
\label{tab:iou}
\begin{tabular}{llcc}
\toprule
Setting & Cameras & IoU & IoU (no walls) \\
\midrule
Cardboard rooms        & 1 & $92.95 \pm 0.05$ & $68.02 \pm 0.54$ \\
Cardboard rooms        & 2 & $95.29 \pm 0.25$ & $77.00 \pm 1.56$ \\
Known photoreal   & 1 & $91.68 \pm 0.12$ & $67.75 \pm 0.38$ \\
Unknown photoreal & 1 & $89.25 \pm 0.55$ & $63.25 \pm 0.53$ \\
Known photoreal   & 2 & $94.54 \pm 0.12$ & $76.87 \pm 0.67$ \\
Unknown photoreal & 2 & $91.87 \pm 0.32$ & $69.96 \pm 0.23$ \\
$12 \times 12$~m room  & 4 & $84.97 \pm 0.33$ & $70.91 \pm 0.60$ \\
\bottomrule
\end{tabular}
\vspace{-19pt}
\end{table}

Table~\ref{tab:iou} reports the results. On known obstacles, moving from cardboard obstacles to photorealistic furniture costs about 1 point of IoU in both configurations, mirroring the small drop in success rate observed in Section~\ref{photorealistic-experiment}. The drop in IoU on held-out furniture assets is larger, mapping to the reported success-rate drop. A second camera raises the interior IoU by roughly 9 points on both cardboard and known photorealistic obstacles, consistent with the two-camera success-rate gain reported in Section~\ref{first-set-of-experiments}. The four-camera $12 \times 12$~m room has the lowest IoU when walls are included because most snapshots contain less wall area. Its interior IoU of $70.91$\% falls between the one- and two-camera cardboard results, providing evidence that the map-estimation module scales to four cameras and a larger room. Figure~\ref{fig:map} shows sampled robot-centric map predictions.

To isolate the contribution of the cross-view transformer, we replace it with a naive ResNet-34 U-Net with a comparable parameter count, trained under the same configuration. In the one-camera \textit{photorealistic rooms} setting, it achieves $83.78 \pm 0.06$\% IoU when walls are counted and $35.73 \pm 0.55$\% when excluded, compared with $91.68 \pm 0.12$\% and $67.75 \pm 0.38$\% for the transformer. In the four-camera $12 \times 12$~m room, it reaches $63.76 \pm 0.14$\% and $29.53 \pm 0.56$\%, against $84.97 \pm 0.33$\% and $70.91 \pm 0.60$\%. The interior IoU drops to roughly half the transformer's value in both settings, indicating that cross-view attention is integral to our method.

\subsection{Real robot experiments}\label{real-robot-experiment}

For real-robot transfer, we retrained both the map-estimation module and the PPO policy in simulation on a \textit{cardboard rooms} variant tuned to include the real setup in its measurements. The real room has black walls, which were not present in previous experiments, so we added a black texture to the pool of wall textures and biased sampling towards it, so that it appears on the walls in $70\%$ of the training episodes. The arena size range is changed to $x \in [2.5, 5.5]$~m and $y \in [2.2, 5.2]$~m, so that the real $4 \times 4$~m room falls inside the training distribution. The obstacle count is $6$ to $14$ per episode when training the policy and $1$ to $12$ when collecting map-estimation snapshots, with footprints in $[0.30, 1.05]$~m by $[0.30, 0.90]$~m and heights in $[0.20, 0.60]$~m. One exocentric camera is used, with a focal length of $16$~mm, a horizontal aperture of $50.2$~mm, and a zoom factor sampled in $[0.58, 1.0]$, covering a field of view wide enough to likely include the real camera. The robot is remodeled to match a Kachaka mobile base, with shell dimensions $0.387 \times 0.240 \times 0.124$~m. The discrete actions apply a forward speed of $0.339$~m/s or a turn of $10.47^\circ$ per step, which at $5$~Hz moves the robot forward by about $6.8$~cm per step, which maps to the real robot where each action is issued as a Kachaka API command of $0.06$~m forward or $10^\circ$ left or right. We used no real images or trajectories during training.

For the map-estimation module, we follow domain randomization~\cite{tobin2017domain} and heavily corrupt the simulated images during training so that they visually approach the real camera feed under the same noise. We blur every image with a Gaussian kernel whose standard deviation is sampled in $[1.2, 1.8]$~px, then add additive pixel noise with standard deviation sampled in $[14, 22]$ on an $8$-bit scale. At deployment, we apply the same blur and noise to the real camera frames, with a fixed blur standard deviation of $1.5$~px and a noise standard deviation of $22$. We also lower the map threshold from $0.5$ to $0.3$ in this setting, so that cells are more likely to be marked as occupied and the policy behaves more conservatively around obstacles. For convenience during the real experiments, we also add a small head that predicts the bearing and normalized distance to the goal from the same image tokens, trained jointly with the map estimation objective. In the real room and simulation, the goal is a green circle placed on the floor.

The real setup is a $4 \times 4$~m room with black walls. We used an Intel RealSense D455 camera, streaming at $1280 \times 800$ pixels, mounted against a wall and pointing into the room. Every camera position reported here was measured with a metric tape and is approximate. The authors manually tuned the camera orientation so the four corners of the room are visible in the frame at every placement.

We evaluate $10$ obstacle layouts and $10$ camera poses, running each layout under each pose for $100$ trials in total. The camera poses combine five positions, at $(0.85, 0.05)$, $(1.80, 0.05)$, $(3.50, 0.05)$, $(4.00, 1.80)$, and $(4.00, 3.70)$~m in room coordinates, with two heights of $2.1$ and $2.3$~m. The five positions span two adjacent walls. Each trial runs until the robot reaches within $0.40$~m of the goal, collides, or exceeds $300$ steps.

ROMUX reaches the goal in $81$ of the $100$ trials, with $8$ collisions and $11$ runs that hit the step cap. Failures relate more to the obstacle layout than to camera placement. Every camera pose succeeds in $7$ to $9$ of its $10$ trials, while the layouts range from all ten trials succeeding to only two. Figure~\ref{fig:real-exp} shows example trials.

We inspected the recorded frames and predicted maps of the failed runs and found three causes. The most common failure mode is the policy stopping in front of narrow gaps between obstacles and turning in place until it reaches the step cap, where it does not commit to passages the predicted map shows as narrow. The other failures come from occlusion of the robot by a tall box, which leaves the map-estimation module without the reference it needs and makes the predicted map unusable for the steps in which the robot is hidden, and the map-estimation module recovering the general arrangement of the obstacles but misplacing their distances and sizes, which puts the robot closer to an obstacle than the map suggests and leads to a collision. We consider this last case acceptable, given that the module was trained only in simulation.

\section{DISCUSSION}\label{limitations}

\textbf{Considerations:} This work aims to show that navigation from unknown and uncalibrated exocentric cameras is possible, rather than to advance navigation performance. We therefore compared against PPO baselines that isolate the effect of the camera configuration, rather than modern navigation agents, which do not operate under this setup. The architecture is also not a deployment-ready system that will work out of the box on any existing environment and camera setup. The perception module was trained entirely in simulation, and generalizing it to arbitrary real environments would likely require more diverse simulation data and possibly fine-tuning on real data.

\textbf{Limitations}: ROMUX assumes that the robot is visible in at least one exocentric image. When it is fully occluded in all views, the model cannot locate it, and the predicted occupancy map becomes useless, which is the most significant limitation of the current architecture. A temporal model that receives a sequence of images and robot actions could estimate the map through occlusion by modeling the expected effect of the actions taken since the robot was last visible. We leave this for future work. A related assumption is that the full room is visible from the set of cameras, so a user would need to position them to cover the entire navigable area.

\section{CONCLUSION}\label{conclusion}

We presented ROMUX, to our knowledge the first agent that navigates a robot toward a point goal while avoiding obstacles using unknown and uncalibrated exocentric cameras. The method can navigate large rooms with partial views, photorealistic environments, and has been transferred zero-shot to a real robot. Future work can improve performance, address limitations and scale to real applications.

\section*{ACKNOWLEDGMENT}
The authors used Claude Opus, versions 4.5 through 5 (Anthropic), for coding and code review, to collect candidate related works that the authors then read, and to improve the writing throughout the manuscript through per-paragraph discussions with the authors, who are not native to english-speaking countries. Grammarly was also used for proofreading.


\bibliographystyle{IEEEtran}
\bibliography{ref}

\end{document}